\documentclass[conference]{IEEEtran}
\IEEEoverridecommandlockouts

\usepackage{cite}
\usepackage{amsmath,amssymb,amsfonts}
\usepackage{algorithmic}
\usepackage{graphicx}
\usepackage{textcomp}
\usepackage{xcolor}
\usepackage{array}
\usepackage{amssymb}
\usepackage{mathrsfs}
\usepackage{algorithm}
\usepackage{booktabs}
\usepackage{multirow}
\usepackage[caption=false,font=normalsize,labelfont=sf,textfont=sf]{subfig}
\usepackage{textcomp}
\usepackage{stfloats}
\usepackage{url}
\usepackage{color}
\usepackage{makecell}
\usepackage{cleveref}
\usepackage{amsthm}
\usepackage{cases}
\theoremstyle{definition}

\usepackage{verbatim}
\usepackage{graphicx}
\usepackage{booktabs}
\usepackage{pifont}
\def\BibTeX{{\rm B\kern-.05em{\sc i\kern-.025em b}\kern-.08em
		T\kern-.1667em\lower.7ex\hbox{E}\kern-.125emX}}
\begin{document}
\newtheorem{theorem}{\bf Theorem}
\newtheorem{lemma}{\bf Lemma}
\newtheorem{corollary}{\it Corollary}
\newtheorem{remark}{\bf Remark}
\newtheorem{example}{\it Example}
\newtheorem{case}{\bf Case Study}
\newtheorem{assumption}{\it Assumption}
\newtheorem{property}{\it Property}
\newtheorem{Corollary}{\bf Corollary}
\newtheorem{proposition}{\bf Proposition}
\renewcommand{\algorithmicrequire}{\textbf{Input:}}
\renewcommand{\algorithmicensure}{\textbf{Output:}}

\newcommand{\hP}[1]{{\boldsymbol h}_{{#1}{\bullet}}}
\newcommand{\hS}[1]{{\boldsymbol h}_{{\bullet}{#1}}}

\newcommand{\ba}{\boldsymbol{a}}
\newcommand{\baq}{\overline{q}}
\newcommand{\bA}{\boldsymbol{A}}
\newcommand{\bb}{\boldsymbol{b}}
\newcommand{\bB}{\boldsymbol{B}}
\newcommand{\bc}{\boldsymbol{c}}
\newcommand{\bp}{\boldsymbol{p}}
\newcommand{\bcO}{\boldsymbol{\cal O}}
\newcommand{\be}{\boldsymbol{e}}
\newcommand{\bh}{\boldsymbol{h}}
\newcommand{\bH}{\boldsymbol{H}}
\newcommand{\bl}{\boldsymbol{l}}
\newcommand{\bm}{\boldsymbol{m}}
\newcommand{\bn}{\boldsymbol{n}}
\newcommand{\bo}{\boldsymbol{o}}
\newcommand{\bO}{\boldsymbol{O}}
\newcommand{\bq}{\boldsymbol{q}}
\newcommand{\br}{\boldsymbol{r}}
\newcommand{\bR}{\boldsymbol{R}}
\newcommand{\bs}{\boldsymbol{s}}
\newcommand{\bS}{\boldsymbol{S}}
\newcommand{\bT}{\boldsymbol{T}}
\newcommand{\bw}{\boldsymbol{w}}

\newcommand{\balpha}{\boldsymbol{\alpha}}
\newcommand{\bbeta}{\boldsymbol{\beta}}
\newcommand{\bomega}{\boldsymbol{\omega}}
\newcommand{\bOmega}{\boldsymbol{\Omega}}
\newcommand{\bphi}{\boldsymbol{\phi}}
\newcommand{\bvarpi}{\boldsymbol{\varpi}}
\newcommand{\bpi}{\boldsymbol{\pi}}
\newcommand{\bxi}{\boldsymbol{\xi}}
\newcommand{\bx}{\boldsymbol{x}}
\newcommand{\by}{\boldsymbol{y}}

\newcommand{\cA}{{\cal A}}
\newcommand{\bcA}{\boldsymbol{\cal A}}
\newcommand{\cB}{{\cal B}}
\newcommand{\cE}{{\cal E}}
\newcommand{\cG}{{\cal G}}
\newcommand{\cH}{{\cal H}}
\newcommand{\bcH}{\boldsymbol {\cal H}}
\newcommand{\cK}{{\cal K}}
\newcommand{\cO}{{\cal O}}
\newcommand{\cR}{{\cal R}}
\newcommand{\cS}{{\cal S}}
\newcommand{\dcS}{\ddot{{\cal S}}}
\newcommand{\ds}{\ddot{{s}}}
\newcommand{\cT}{{\cal T}}
\newcommand{\cU}{{\cal U}}
\newcommand{\wt}[1]{\widetilde{#1}}

\newcommand{\mA}{\mathbb{A}}
\newcommand{\mE}{\mathbb{E}}
\newcommand{\mG}{\mathbb{G}}
\newcommand{\mR}{\mathbb{R}}
\newcommand{\mS}{\mathbb{S}}
\newcommand{\mU}{\mathbb{U}}
\newcommand{\mV}{\mathbb{V}}
\newcommand{\mW}{\mathbb{W}}

\newcommand{\uq}{\underline{q}}
\newcommand{\ubq}{\underline{\boldsymbol q}}

\newcommand{\red}[1]{\textcolor[rgb]{1,0,0}{#1}}
\newcommand{\gre}[1]{\textcolor[rgb]{0,1,0}{#1}}
\newcommand{\blu}[1]{\textcolor[rgb]{0,0,1}{#1}}

\title{Robust Federated Learning against Model Perturbation in Edge Networks
}

\author{\IEEEauthorblockA{Dongzi~Jin\IEEEauthorrefmark{1}, Yong~Xiao\IEEEauthorrefmark{1}\IEEEauthorrefmark{3}, Yingyu Li\IEEEauthorrefmark{2} \\
		\IEEEauthorblockA{\IEEEauthorrefmark{1}School of Electronic Inform. \& Commun., Huazhong University of Science \& Technology, China}
		\IEEEauthorblockA{\IEEEauthorrefmark{2}School of Mech. Eng. and Elect. Inform., China University of Geosciences, China}
		\IEEEauthorblockA{\IEEEauthorrefmark{3}Pengcheng Lab, Guangzhou, China}
	}
}

\maketitle

\begin{abstract}
	Federated Learning (FL) is a promising paradigm for realizing edge intelligence, allowing collaborative learning among distributed edge devices by sharing models instead of raw data. However, the shared models are often assumed to be ideal, which would be inevitably violated in practice due to various perturbations, leading to significant performance degradation. To overcome this challenge, we propose a novel method, termed Sharpness-Aware Minimization-based Robust Federated Learning (SMRFL), which aims to improve model robustness against perturbations by exploring the geometrical property of the model landscape. Specifically, SMRFL solves a min-max optimization problem that promotes model convergence towards a flat minimum by minimizing the maximum loss within a neighborhood of the model parameters. In this way, model sensitivity to perturbations is reduced, and robustness is enhanced since models in the neighborhood of the flat minimum also enjoy low loss values. The theoretical result proves that SMRFL can converge at the same rate as FL without perturbations. Extensive experimental results show that SMRFL significantly enhances robustness against perturbations compared to three baseline methods on two real-world datasets under three perturbation scenarios.
\end{abstract}

\vspace{-0.1cm}
\section{Introduction}
Federated Learning (FL) is an emerging machine learning paradigm that aligns well with the distributed nature of edge networks\cite{XY2024TMCTSFL}. It enables multiple edge devices to learn a global model collaboratively by transmitting local models instead of local data, thereby mitigating privacy concerns and reducing communication overhead caused by centralized data collection\cite{XY2024TMCTraffSynth, jdz_vtc}.
Many efforts have been made to further improve communication efficiency, resource management, and privacy in FL-enabled edge networks. 

However, these existing works often assume ideal model sharing over edge networks, overlooking the reality of model perturbations. In practice, received models are inevitably lossy, stemming from three primary sources of perturbation: (\romannumeral1) the communication environment, where factors like channel noise, channel fading, and inter-cell interference can distort the received model; (\romannumeral2) the system designs, many of which, though initially intended to improve system performance, can unintentionally introduce perturbations as side effects, as exemplified by weight quantization, and differential privacy (DP) schemes; and (\romannumeral3) the adversarial attacks, where malicious clients purposely employ methods such as data poisoning and model corruption to compromise the learned model. In the following context, we refer to perturbations from the third source as malicious perturbations, while we refer to perturbations from the first two sources as non-malicious perturbations. 

This work focuses on the problem setting of mitigating the non-malicious perturbations. It is worth noting that even non-malicious perturbations on the model require careful consideration; otherwise, they can severely degrade the performance and learning efficiency of FL methods. In \cite{robust_qi_2022}, it was demonstrated that additive random Gaussian noise of 3-5 dB can lead to about 10\% of test accuracy degradation. Similarly, results in\cite{sparse_dp_hu} verified that privacy guarantees provided by the DP scheme could come at the cost of accuracy degradation due to operations of gradient clipping and adding artificial noise.

Current research in this field generally takes two primary approaches. One line of approach to address non-malicious perturbations is to reuse the defenses against malicious perturbations. However, defenses for malicious perturbation are generally unsuitable and can’t be directly applied in this problem setting for three main reasons. First, computation and communication costs for defenses specialized for malicious perturbations may lead to a huge burden on the edge networks and compromise the learning efficiency\cite{compu_expensive_1}. Second, the perturbation modes are generally different. Typically, malicious perturbations are bursty or periodic\cite{NEURIPS2021_692baebe}, while non-malicious perturbations persist during the whole process of model training and affect nearly all clients. Last, non-IID (Independently and Identically Distributed) data can undermine the effectiveness of adversarial defenses by making it difficult to differentiate between perturbed and unperturbed models\cite{zhao2022fedinv}. 
The other line of approach is to develop countermeasures specifically designed for non-malicious perturbations, which have fewer existing works. In \cite{qua2022honig}, authors proposed adapting the quantization level to alleviate the performance degradation sourced from model quantization. In\cite{sparse_dp_hu}, the DP noise was added to a selected subset of coordinates to mitigate the negative impact on accuracy sourced from DP noise. Methods in \cite{qua2022honig} and \cite{sparse_dp_hu} were limited to the scenario where perturbations were only involved in the uplink of model sharing. In contrast, both uplink and downlink perturbations sourced from channel noise were considered in \cite{wei2022noise}. However, the transmit power control scheme needs to scale the signal-to-noise ratio as $\mathcal{O}(r^2)$ to provide the same convergence rate of FL without noise and perfect channel state information (CSI) was assumed in\cite{wei2022noise}, where $r$ is the index of communication rounds. Therefore, how to design a robust FL method against non-malicious perturbations with wider scenarios and more practical assumptions is still under-explored.

\begin{figure}[htbp]
	\vspace{-0.3cm}
	\centering
	\includegraphics[width=0.35\textwidth]{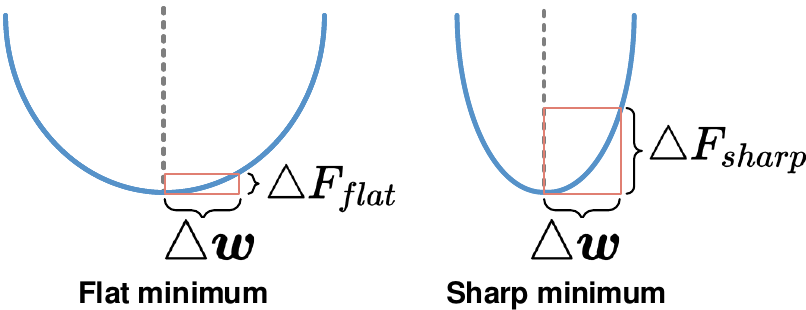}
	\vspace{-0.2cm}						
	\caption{Flat minimum v.s. sharp minimum (Blue lines are 2-dimension loss landscape.  Under the same strength of model perturbation $\triangle \boldsymbol{w}$, loss value of the flat minimum has smaller changes ($\triangle F_{flat} < \triangle F_{fsharp}$).}
	\label{sharp_flat}
	\vspace{-0.3cm}
\end{figure}

This motivates us to proposes a Sharpness-aware Minimization-based Robust FL (SMRFL) method to handle a wide source of non-malicious perturbations with both uplink and downlink scenarios in a unified framework. To achieve this goal, SMRFL explores the geometrical property of flat minima in the loss landscape. Flat minima are model points where the loss varies slowly in the neighborhood, as shown in Fig. \ref{sharp_flat}. This kind of minimum is less sensitive to model changes and enjoys the advantage of robustness against perturbations since similar small loss values are maintained  over its neighborhood in the model space. This geometrical property make the model itself robust\cite{foret2020sharpness}, and therefore alleviating the need of stringent assumptions in existing works. To obtain flatter minima, SMRFL performs local model updates using sharpness-aware minimization (SAM) instead of empirical risk minimization (ERM) as in traditional FL methods. 
The main contributions are summarized as follows:
\begin{itemize}
	\item To cope with the model performance degradation caused by non-malicious perturbations, we propose a novel method termed SMRFL, which encourages convergence to flatter minima, reducing sensitivity to model changes and enhancing robustness against variations in model parameter.
	\item We present the theoretical results characterizing the convergence rate of SMRFL with general non-convex loss functions. SMRFL algorithm can achieve a convergence rate of $\mathcal{O}(\frac{1}{\sqrt{R}})$, matching the convergence rate of FL without perturbations, where $R$ is the total communication rounds.
	\item We provide extensive experimental results on two real-world datasets with both IID and non-IID data settings. Our results validate that SMRFL can achieve more robust model performance against non-malicious perturbations with both uplink and downlink scenarios and efficiently converge to flatter minima, compared with three baseline methods.
\end{itemize}
%
%

\section{System Model and Problem Formulation} \label{sec_sys}
In this section, we will first describe the system model and then introduce FL with model perturbations. Finally, the optimization problem is formulated.
	

\subsection{System Model}
\begin{figure}[htbp]
	\centering
	\includegraphics[width=3.5in]{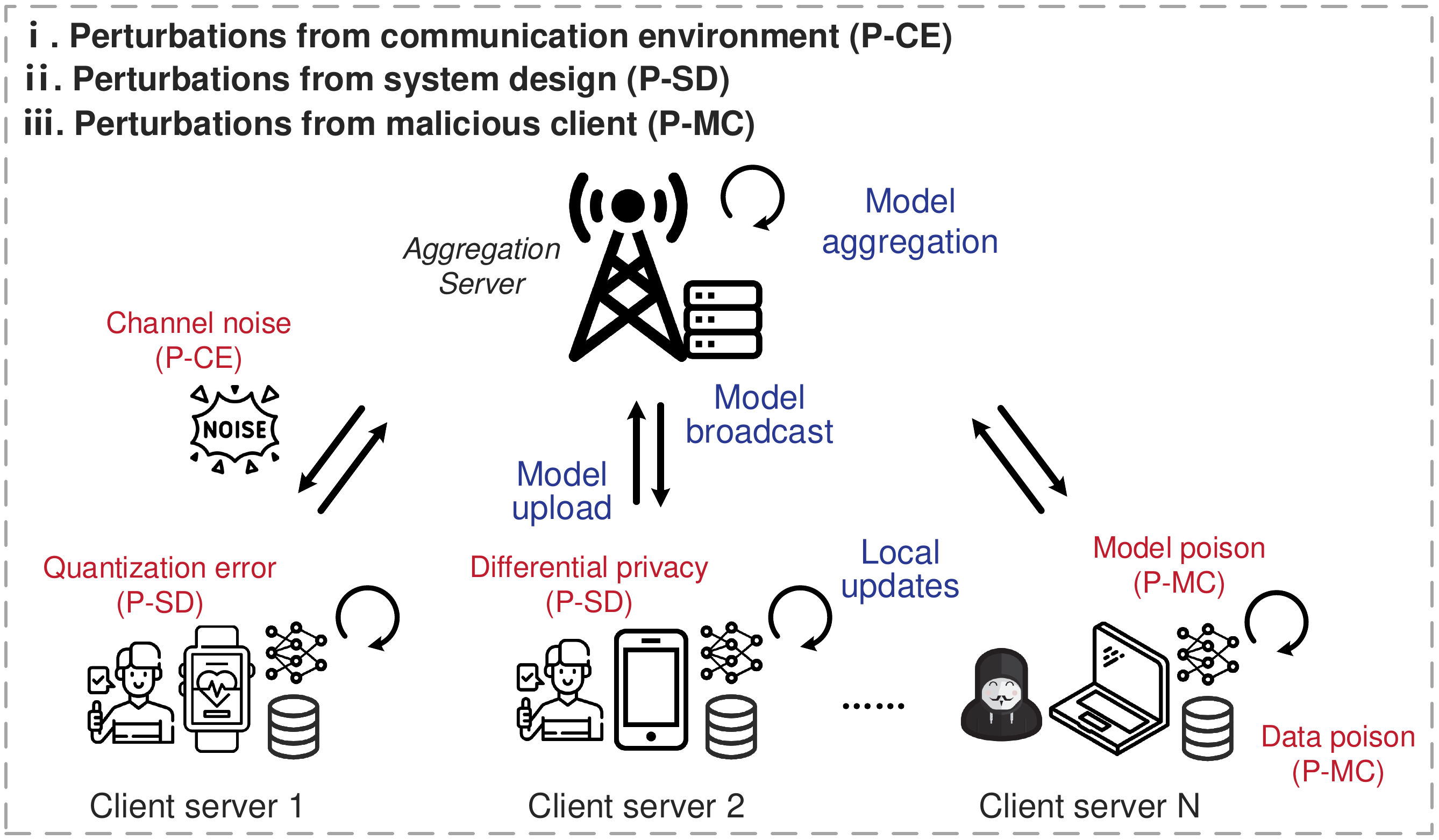}
	\vspace{-0.6cm}
	\caption{System model of robust FL (FL steps in blue font and typical examples of perturbation in red font)}
	\label{system_model}
	\vspace{-0.4cm}
\end {figure}

As shown in Fig. \ref{system_model}, we consider a FL system over edge network consisting of an aggregation server and a set of client servers $\mathcal{N}=\{1,\ldots, N\}$. Each client server possesses a local model $\boldsymbol{w}^i \in \mathbb{R}^d$ and a private local dataset $\mathcal{D}_i$ with $D_i$ data samples. The datasets are stored locally at the client servers. The local loss function of a client server is defined as $F_i(\boldsymbol{w}^i)=\mathbb{E}_{\xi^i \sim \mathcal{D}_i}F_i(\boldsymbol{w}^i,\xi^i)$ to measure the expected loss of the local model $\boldsymbol{w}_i$ over data sample $\xi^i$ from local dataset $\mathcal{D}_i$, where $F_i(\boldsymbol{w}^i,\xi^i)$ represents the sample-wise loss function, e.g., mean square error and cross-entropy. In practice, the loss is approximated by the empirical loss $\hat{F}_i(\boldsymbol{w}^i)=\frac{1}{D_i}\sum_{\xi_i \in \mathcal{D}_i}F(\boldsymbol{w}^i,\xi^i)$.

\vspace{-0.3cm}

\subsection{FL with Model Perturbations} \label{perturbation_discussion}
The perturbed model $\hat{\boldsymbol{w}}$ is formulated by involving a perturbation term $\boldsymbol{\epsilon}$ to represent the negative impacts from the communication environment and side effects of system design.
\vspace{-0.5cm}
\begin{align}
	\hat{\boldsymbol{w}} = \boldsymbol{w}+\boldsymbol{\epsilon},
\end{align}
where $l_2$-norm of $\boldsymbol{\epsilon}$ is bounded by $\rho$ ($\|\boldsymbol{\epsilon}\|_2 \leq \rho$). This means $\hat{\boldsymbol{w}}$ can be understood as a point in the parameter space within the $\rho$-ball center at the original model $\boldsymbol{w}$. This perturbed form is general since we only bound the amplitude of the perturbation, and the distribution of the perturbation is not specified. Moreover, this additive formulation of the perturbed model can cover many sources of perturbations. For example, in \cite{Amiri2022_noise}, the channel noise is a vector following a circularly symmetric complex Gaussian distribution adding to the model. Likewise, local DP is achieved by adding Gaussian noise to the gradients before uploading\cite{sparse_dp_hu}. Moreover, the original parameters can also be expressed in an additive form of quantized parameters and the quantization error\cite{qua2022honig, qua2021shle}. 

Then, we will introduce the operation steps of  FL with model perturbations. At the $r$th communication round, the operational steps can be described as:

\noindent\textbf{Perturbed model broadcasting:} At the beginning of the $r$th communication round, the aggregation server broadcasts the global model $\boldsymbol{w}_r$ to the clients. Due to the impact of perturbations, the client server $i$ receives a perturbed version of $\boldsymbol{w}_r$, which is given as
\vspace{-0.2cm}
\begin{align}
	\hat{\boldsymbol{w}}_r^i = \boldsymbol{w}_r + \boldsymbol{\epsilon}_{down,r}^i,
\end{align}
where $\boldsymbol{\epsilon}_{down,r}^i \in \mathbb{R}^d$ is the downlink perturbation vector at client $i$.  

\noindent\textbf{Local model update:} Client $i$ initializes the local model with the received global model and performs local model updates by sampling data $\xi_r^i$ from local dataset $\mathcal{D}_i$. After $E$ steps of updates, local model $\boldsymbol{w}^i_{t,E}$ is uploaded to the aggregation server. The local model update operations of client $i$ are given as
\vspace{-0.2cm}
\begin{align}
	&\boldsymbol{w}^i_{r,0} = \hat{\boldsymbol{w}}_{r-1}^i \quad \quad\quad\quad\quad\quad\quad\quad\quad(initialization) \\				
	&\boldsymbol{w}^i_{r,\tau} = \boldsymbol{w}^i_{r,\tau-1} - \eta_l\nabla F_i(\boldsymbol{w}^i_{r,\tau-1},\xi_{\tau-1}^i)\quad(update) \label{standard_update}\\
	& \boldsymbol{w}^i_r = \boldsymbol{w}^i_{r,E} \quad\quad\quad\quad\quad\quad\quad\quad\quad\quad\quad\quad\quad(upload)
\end{align}
\textbf{Perturbed local model uploading:} The local models received by the aggregation server could also be perturbed. The received local model from client $i$ is given as
\begin{align}
	\hat{\boldsymbol{w}}_r^i = \boldsymbol{w}^i_r + \boldsymbol{\epsilon}_{up,r}^i,
\end{align}
where $\boldsymbol{\epsilon}_{up,r}^i \in \mathbb{R}^d$ is the uplink perturbation vector of the local model from client $i$. 

\noindent\textbf{Global model aggregation:} The aggregation server aggregates the perturbed local models to obtain an updated global model by
\vspace{-0.3cm}
\begin{align}
	\boldsymbol{w}_r = \frac{1}{N} \sum_{i \in \mathcal{N}} \hat{\boldsymbol{w}}_r^i,
\end{align}
where perturbed local models have the same aggregation weight of $\frac{1}{N}$ for simplification. The weight here can be easily generalized to the case of adaptive weights. 

The above steps repeat until the model converges or a predefined total communication rounds $R$ are reached.

\subsection{Problem formulation}
We briefly review the optimization problem in the standard FL setting, which is given as
\vspace{-0.2cm}
\begin{align}
	\min_{\boldsymbol{w}}\{F(\boldsymbol{w})=\frac{1}{N}\sum_{i \in \mathcal{N}}F_i(\boldsymbol{w})\}, \label{fl_obj}
\end{align}
where $ F_i(\boldsymbol{w})$ is the local loss function over unperturbed model $\boldsymbol{w}$. This formulation contains an implicit assumption that the received models at both the aggregation server (uplink) and the client servers (downlink) are unperturbed.

In contrast, this work studies a more practical setting, where the received models are lossy due to non-malicious perturbations. 
To achieve the goal of robustness, the optimization problem is designed following two principles: first, the global loss should be minimized to achieve a good model performance; second, the sharpness of the obtained model should be minimized, which means the model is robust against perturbations and maintains a low loss even when the model is perturbed. Under the above principles, the optimization problem is formulated as:
\vspace{-0.1cm}
\begin{align}
	\mathcal{P:} \quad \min_{\boldsymbol{w}}\max_{ \|\boldsymbol{\delta_i}\|_2 \leq \rho} \{F(\hat{\boldsymbol{w}})=\frac{1}{N} \sum_{i \in \mathcal{N}} F_i(\hat{\boldsymbol{w}})\}, \label{rfl_obj}
\end{align}
where $\boldsymbol{\delta_i}$ is the effective perturbation vector to represent perturbations from both uplink and downlink, $\hat{\boldsymbol{w}} = \boldsymbol{w} + \boldsymbol{\delta_i}$ is the perturbed model, and $F_i(\hat{\boldsymbol{w}}) = F_i(\boldsymbol{w}+\boldsymbol{\delta_i})$ is the expected loss over perturbed model. The inner loop of the objective function is to find a worst-case model that can maximize the global loss in $\rho$-neighborhood, while the outer loop finds a model that can minimize the maximized global loss under perturbation. By rewrite the problem $\mathcal{P}$ in the following form:
\vspace{-0.1cm}
\begin{align}
	\min_{\boldsymbol{w}}\underbrace{\max_{ \|\boldsymbol{\delta_i}\|_2 \leq \rho} \{\frac{1}{N} \sum_{i \in \mathcal{N}} F_i(\hat{\boldsymbol{w}})\}-\frac{1}{N} \sum_{i \in \mathcal{N}} F_i(\boldsymbol{w})}_{\mathcal{A}_1}+\underbrace{\frac{1}{N} \sum_{i \in \mathcal{N}} F_i(\boldsymbol{w})}_{\mathcal{A}_2},
\end{align}
where it's easy to see that the objective of problem $\mathcal{P}$ is following the principles discussed above by simultaneously minimizing sharpness (term $\mathcal{A}_1$) and global loss (term $\mathcal{A}_2$). 	
The difficulty of solving problem $\mathcal{P}$ lies in the complexity of solving the min-max optimization problem locally at the client servers during each communication round. The non-convexity of loss function further exacerbates this situation.

\section{The Proposed Robust FL Method}  
\label{sec_SMRFL}

We design a Sharpness-aware Minimization-based Robust FL (SMRFL) method by addressing the challenging distributed min-max optimization problem in (\ref{rfl_obj}). In each communication round, finding an exact solution for the maximization is computationally expensive for the client servers. Therefore, we seek to transform the problem $\mathcal{P}$ by approximating the maximization problem in the inner loop by using the first-order Taylor expansion of $F_i(\boldsymbol{w}+\boldsymbol{\delta}^i)$:
\vspace{-0.1cm}
\begin{align}
	\boldsymbol{\delta}^i = & \arg \max_{\|\boldsymbol{\delta}^i\| \leq \rho} F_i(\boldsymbol{w}+\boldsymbol{\delta}^i) \\
	\approx & \arg \max_{\|\boldsymbol{\delta}^i\| \leq \rho} \{F_i(\boldsymbol{w})+{\boldsymbol{\delta}^i}^T \nabla F_i(\boldsymbol{w}) \} \\
	= & \rho \frac{\nabla F_i(\boldsymbol{w})}{\|\nabla F_i(\boldsymbol{w})\|}.
\end{align}

Therefore, the transformed problem $\mathcal{P}_1$ is given as
\begin{align}
	\mathcal{P}_1: \quad \min_{\boldsymbol{w}} \{F(\boldsymbol{w})=\frac{1}{N} \sum_{i \in \mathcal{N}} F_i(\boldsymbol{w} + \rho\frac{\nabla F_i(\boldsymbol{w})}{\|\nabla F_i(\boldsymbol{w})\|}) \}. \label{p1_obj}
\end{align} 

\begin{figure}[htbp]
\vspace{-0.3cm}
\centering
\includegraphics[width=2in]{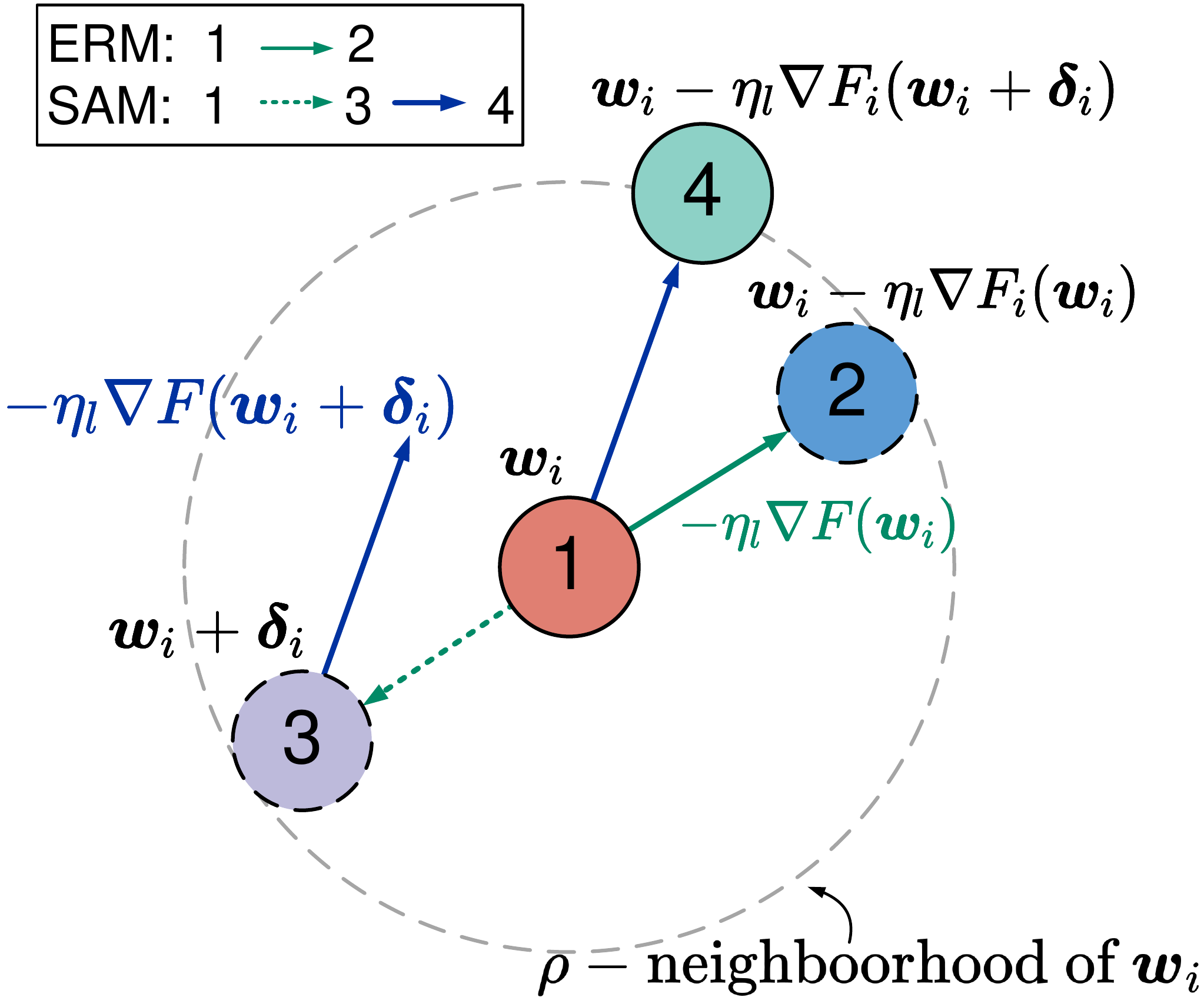}
\caption{Difference between SAM and ERM in local (the plane is the 2-dimensional model space, and colored circles are models in the space)}
\label{sam_vs_erm}
\vspace{-0.3cm}
\end {figure}

We can observe that the problem of $\mathcal{P}_1$ in (\ref{p1_obj}) belongs to the finite-sum optimization problem as that of the standard FL in (\ref{fl_obj}). Therefore, $\mathcal{P}_1$ can be solved following the typical process of FL, except that the local model update of SMRFL is modified into the following two steps (as in line 7-8 of Algorithm \ref{rfl_alg_1}):
\begin{numcases}{}
	\hat{\boldsymbol{w}}_{r,\tau}^i = \boldsymbol{w}_{r,\tau}^i + \rho\frac{\nabla F_i(\boldsymbol{w}_{r,\tau}^i,\xi_{r,\tau}^i)}{\|\nabla F_i(\boldsymbol{w}_{r,\tau}^i,\xi_{r,\tau}^i)\|},  \label{local_perturbation}\\
	\boldsymbol{w}_{r,\tau+1}^i = \boldsymbol{w}_{r,\tau}^i -\eta_l \nabla F(\hat{\boldsymbol{w}}_{r,\tau}^i), \label{update_after_perturbation}
\end{numcases}
where the worst-case model in the $\rho$-neighborhood of the current local model $\boldsymbol{w}_{r,\tau}^i$ is obtained as in (\ref{local_perturbation}); then local model is updated using gradient computed on the worst-case model in the $\rho$-neighborhood as in (\ref{update_after_perturbation}). To make the illustration clear, the difference in model update process between standard FL using empirical risk minimization (ERM) as in (\ref{standard_update}) and SMRFL using SAM as in (\ref{local_perturbation}) and (\ref{update_after_perturbation}) is shown in Fig. \ref{sam_vs_erm}. SAM achieves flat minima by first finding the model with the largest loss value (worst-case model) in the neighborhood of the current model and then optimizing the model using the gradient of the worst-case model so that the maximum loss within the neighborhood of the current model is minimized\cite{foret2020sharpness}. The proposed SMRFL is detailed in Algorithm \ref{rfl_alg_1}.
	
\begin{algorithm}
\caption{SMRFL}
\label{rfl_alg_1}
\begin{small}
\begin{algorithmic}[1]
	\REQUIRE{initial model parameters $\boldsymbol{w}_0$, $\forall i \in \mathcal{N}$; step sizes $\eta_l$, $\eta_g$; synchronization gap $E$; communication rounds $R$}
	\ENSURE{Global model $\boldsymbol{w}_R$}
	
	\FOR{$r=0, 1, \ldots, R-1$}
	\STATE{Aggregation server broadcasts the global model $\boldsymbol{w}_r$ to the clients;}
	\FOR{each client $i \in \mathcal{N}$ in parallel}
	\STATE {$\boldsymbol{w}^i_{r,0} = \boldsymbol{w}_r$;}
	\FOR{$\tau=0, 1, \ldots, E-1$}
	\STATE {Client $i$ randomly samples a data sample $\xi_{r,\tau}^i$ from local dataset $\mathcal{D}_i$ and computes the gradient: \\
		$\nabla F_i\left(\boldsymbol{w}_{r,\tau}^i, \xi_{r,\tau}^i\right)$; 
	}
	\STATE {Compute the perturbed model: \\
		$\hat{\boldsymbol{w}}_{r,\tau}^i= \boldsymbol{w}_{r,\tau}^i + \rho \frac{\nabla F_i\left(\boldsymbol{w}_{r,\tau}^i, \xi_{r,\tau}^i\right)}{\|\nabla F_i\left(\boldsymbol{w}_{r,\tau}^i, \xi_{r,\tau}^i\right)\|_2}$ ;}
	\STATE{The local model is updated as: \\
		$\boldsymbol{w}^i_{r,\tau+1} = \boldsymbol{w}^i_{r,\tau} - \eta_l \nabla F_i\left(\hat{\boldsymbol{w}}_{r,\tau}^i, \xi_{r,\tau}^i\right);$ 
	}
	\ENDFOR
	\STATE{Clients update the local model $\boldsymbol{w}^i_r = \boldsymbol{w}^i_{r,E}$ to the aggregation server;} 
	\ENDFOR
	\STATE{Central server updates the global model as: \\
		$\boldsymbol{w}_{r+1} = \boldsymbol{w}_{r} + \eta_g \frac{1}{N}\sum_{i\in \mathcal{N}}(\boldsymbol{w}^i_r-\boldsymbol{w}_{r})$;
	}
	\ENDFOR
\end{algorithmic}
\end{small}

\end{algorithm}
\addtolength{\topmargin}{-.4cm}				
\vspace{-0.3cm}
\section{Convergence Analysis of SMRFL}
\label{theo_result}
In this section, we will provide the convergence analysis of SMRFL with general non-convex loss functions. 

Before that, we introduce the following assumptions:
%
	%
%
\vspace{-0.2cm}
\begin{assumption}
	(Smoothness). Function $F_i(\boldsymbol{w})$ is $L$-smooth for all $i \in \mathcal{N}$, i.e., $\|\nabla F_i(\boldsymbol{w})-\nabla F_i(\boldsymbol{v})\| \leq L\|\boldsymbol{w}-\boldsymbol{v}\|$, for all $\boldsymbol{w},\boldsymbol{v} \in \mathbb{R}^d$ and $i \in \mathcal{N}$.
	\label{assump_smooth}
	\vspace{-0.2cm}
\end{assumption}
\begin{assumption}
	(Bounded Variance of Local Gradient). The variance of local gradients with respect to the global gradient is bounded by $\sigma_g^2$ for all clients, i.e., $\|\nabla F_i(\boldsymbol{w}_r)-\nabla F(\boldsymbol{w}_r)\|^2 \leq \sigma_g^2, \forall i \in \mathcal{N}$.
	\label{assump_heterogeneity}
	\vspace{-0.2cm}
\end{assumption}
\begin{assumption}
	(Bounded Variance of Stochastic Gradient). The stochastic gradient $\nabla F_i(\boldsymbol{w}, \xi^{i})$ is an unbiased estimator of the full gradient$\nabla F_i(\boldsymbol{w})$, and its variance is bounded by $\sigma_l^2$ for all clients, i.e., $\mathbb{E}[\nabla F_i(\boldsymbol{w}, \xi^{i})] = \nabla F_i(\boldsymbol{w})$, and $\mathbb{E}\|\frac{\nabla F_i(\boldsymbol{w}, \xi^{i})}{\|\nabla F_i(\boldsymbol{w}, \xi^{i})\|}-\frac{\nabla F_i(\boldsymbol{w})}{\|\nabla F_i(\boldsymbol{w})\|}\|^2\leq \sigma_l^2, \forall i \in \mathcal{N}$.
	\label{assump_stochastic}
\end{assumption}
\vspace{-0.2cm}
\begin{remark}
	Assumptions \ref{assump_smooth}, and \ref{assump_heterogeneity} are widely used in general non-convex FL context\cite{scaffold}. Assumption \ref{assump_stochastic} is from \cite{fedsam}, which is tighter than the similar assumption $\mathbb{E}\|\nabla F_i(\boldsymbol{w}, \xi^{i})-\nabla F_i(\boldsymbol{w})\|^2\leq \sigma_l^2, \forall i \in \mathcal{N}$ used in \cite{scaffold}. Assumption \ref{assump_stochastic} indicates that stochastic gradient descent is used, which can be easily extended to mini-batch gradient descent by assuming $\mathbb{E}\|\frac{\nabla F_i(\boldsymbol{w}, \xi^{i})}{\|\nabla F_i(\boldsymbol{w}, \xi^{i})\|}-\frac{\nabla F_i(\boldsymbol{w})}{\|\nabla F_i(\boldsymbol{w})\|}\|^2\leq \frac{\sigma_l^2}{b}, \forall i \in \mathcal{N}$, where $b$ is the batch size of data samples. 
\end{remark}
\vspace{-0.2cm}
\begin{theorem} 
	Suppose the assumptions \ref{assump_smooth}-\ref{assump_stochastic} hold. There exists $\eta_g\eta_l \leq \frac{1}{2EL}$ and $\eta_l < \frac{1}{4EL} $ the convergence rate of SMRFL in algorithm \ref{rfl_alg_1} after $R$ communication rounds is given by
	\begin{align}
		\frac{1}{R}\sum_{r=0}^{R-1} \mathbb{E} \|\nabla F(\boldsymbol{w}_r)\|^2 \leq & \frac{\nabla F(\hat{\boldsymbol{w}}_0)-\nabla F(\hat{\boldsymbol{w}}_*)}{C E \eta_g \eta_l R} + \Omega, \label{convergence_no_VR_eq}
	\end{align}
	where $\Omega = \frac{1}{CE\eta_g\eta_l}[8eE^3L^4\eta_g\eta_l^3\rho^2\sigma_l^2 + 24eE^3L^2\eta_g\eta_l^3\sigma_g^2 + 32eE^5L^6\eta_l^5 \rho^2 + (48e+2)E^3L^4\eta_g\eta_l^3\rho^2 + EL^3\rho^2\sigma_l^2 \frac{\eta_g^2 \eta_l^2}{N}]$ and $C$ is a constant that satisfies $\frac{1}{2}-8eE^2L^2\eta_l^2 > C > 0 $.	
	\label{convergence_no_VR}
\end{theorem}
\vspace{-0.4cm}
\begin{remark}
	(Proof Outline) The proof of theorem \ref{convergence_no_VR} proceeds as follows: ($\romannumeral1$) the bound of one-step progress $F(\boldsymbol{w}_{r+1})-F(\boldsymbol{w}_{r})$ is built; ($\romannumeral2$) the client drift is bounded; ($\romannumeral3$) the proof is concluded by substituting the bound of client drift into one-step progress, and summing one-step progress over all communication rounds. Due to limited space, only the proof outline is provided here.
\end{remark}
\vspace{-0.4cm}
\begin{remark}
	The convergence rate comprises two parts in equation (\ref{convergence_no_VR_eq}). The first part quantifies the error induced by the model initialization, which vanishes as $R$ increases. The second part is a constant $\Omega$ determined by the problem parameters. The first four terms in $\Omega$ (i.e.,$\frac{\eta_l^2}{CE}[8eE^3L^4\rho^2\sigma_l^2 + 24eE^3L^2\sigma_g^2 + 32eE^5L^6\eta_l^5 \rho^2 + (48e+2)E^3L^4\rho^2]$) are independent of N and are determined by variance of local gradient, variance of stochastic gradient, and the perturbation amplitude. To reduce the first four terms, a small local learning rate $\eta_l = \mathcal{O}(\frac{1}{E})$ can be chosen. The last term in $\Omega$ (i.e.,$L^3\rho^2\sigma_l^2 \frac{\eta_g \eta_l}{CN}$) is decreasing with the number of clients $N$.  
\end{remark}
\vspace{-0.4cm}
\begin{remark}
	\label{corollary_1}
	Choosing the learning rates $\eta_l = \mathcal{O}(\frac{1}{\sqrt{R}EL})$, $\eta_g = \sqrt{EN}$, and the neighborhood radius $\rho = \mathcal{O}(\frac{1}{\sqrt{R}})$ in theorem \ref{convergence_no_VR}, we have
	\vspace{-0.2cm}
	\begin{align}
		& \frac{1}{R}\sum_{r=0}^{R-1} \mathbb{E} \|\nabla F(\boldsymbol{w}_r)\|^2 \nonumber \\
		= & \mathcal{O}(\frac{\nabla F(\hat{\boldsymbol{w}}_0)-\nabla F(\hat{\boldsymbol{w}}_*)}{\sqrt{REN}}+\frac{\sigma_g^2}{R} + \frac{L^2 \sigma_l^2}{R^{3/2}\sqrt{EN}}),
	\end{align}
	where the higher-order terms are omitted. 
\end{remark}
\vspace{-0.2cm}
From remark \ref{corollary_1}, we observe that by choosing learning rates $\eta_l$ and $\eta_g$ the convergence rate is $\mathcal{O}(\frac{1}{\sqrt{R}})$ when the communication round $R$ is large, which matches the convergence rate of FL with general non-convex loss function and without perturbations.

%
%
			
\begin{figure*}[htbp]
	\centering
	\begin{subfloat}[\footnotesize Downlink perturbation]
		{
			\includegraphics[width=0.26\textwidth]{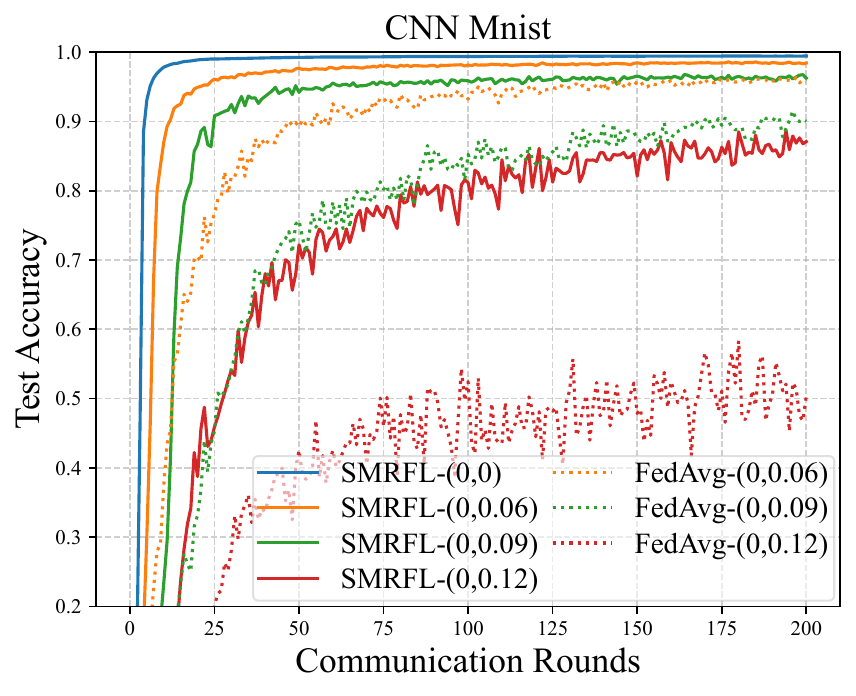}				
			\label{exp1_1}
		}
	\end{subfloat}
	\hspace{-0.5cm} \vspace{-0.1cm}
	\begin{subfloat}[\footnotesize Uplink perturbation]{
			\includegraphics[width=0.27\textwidth]{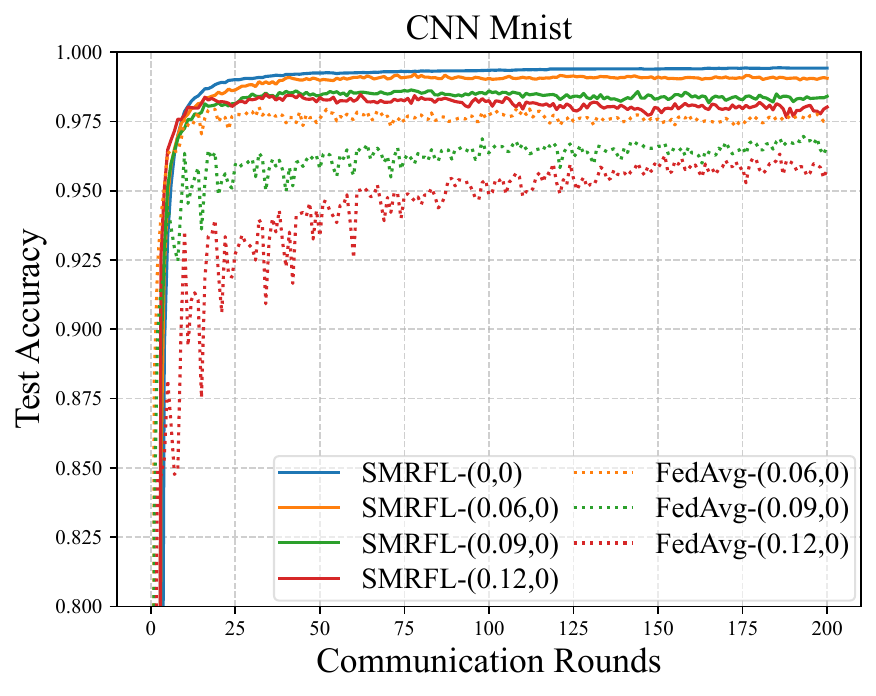}				
			\label{exp1_2}
		}				
	\end{subfloat}
	\hspace{-0.5cm} \vspace{-0.1cm}
	\begin{subfloat}[\footnotesize Uplink and downlink perturbation]{
			\includegraphics[width=0.26\textwidth]{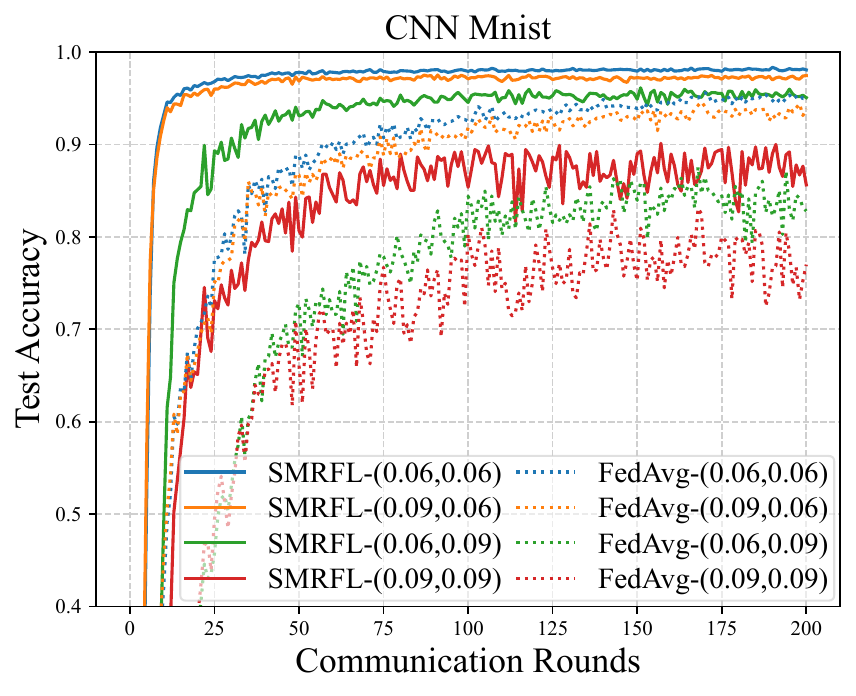}		
			\label{exp1_3}
		}				
	\end{subfloat} \hspace{-0.5cm} \vspace{-0.1cm}
	\begin{subfloat}[\footnotesize Downlink perturbation]{
			\includegraphics[width=0.26\textwidth]{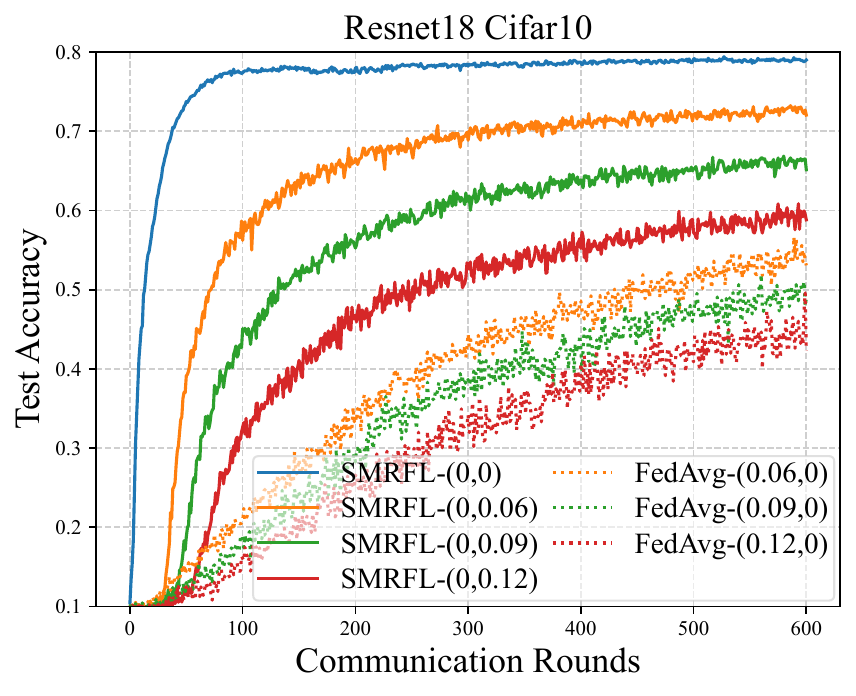}		
			\label{exp1_4}
		}				
	\end{subfloat}
	\hspace{-0.5cm} 
	\begin{subfloat}[\footnotesize Uplink perturbation]{
			\includegraphics[width=0.26\textwidth]{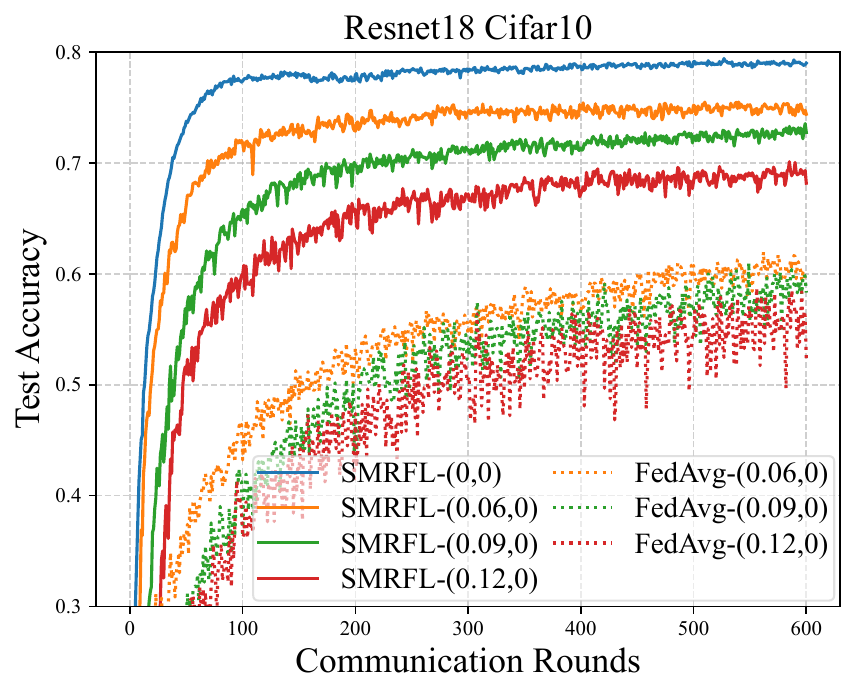}		
			\label{exp1_5}
		}				
	\end{subfloat}
	\hspace{-0.5cm}
	\begin{subfloat}[\footnotesize Uplink and downlink perturbation]{
			\includegraphics[width=0.26\textwidth]{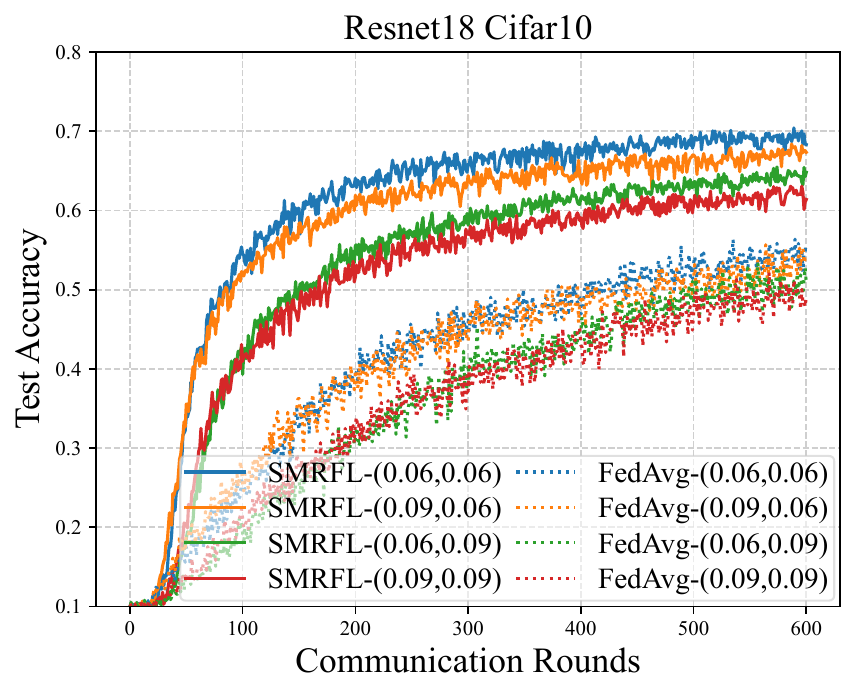}		
			\label{exp1_6}
		}				
	\end{subfloat}
	
	\caption{Test accuracy of SMRFL compared with FedAvg under three scenarios: only downlink perturbation, only uplink perturbation, and both uplink and downlink perturbations, where $(x,y)$ in the legend means uplink perturbation strength of $x$ and downlink perturbation strength of $y$.}
	\label{exp1}
	\vspace{-0.6cm}
\end{figure*}

\vspace{-0.3cm}			
\section{Performance evaluation} \label{sec_exp}

\subsection{Experimental Setup}

\textbf{1) Dataset and Data Partition:}
The performance of the proposed SMRFL is evaluated on Mnist and Cifar-10 datasets. Data samples are partitioned into both IID and non-IID settings, and the non-IID degree is controlled by $\alpha$ following the Dirichlet distribution, where smaller $\alpha$ means a larger non-IID degree.
\textbf{2) Model Architecture and Hyperparameter:}
Simple CNN and Resnet18 are used for Mnist and Cifar-10, respectively. The simple CNN model consists of two convolutional layers with 5x5 kernels, each followed by ReLU activation and 2x2 max-pooling, and two fully connected layers; ResNet18 consists of an initial 7x7 convolutional layer followed by a max-pooling layer, then four residual blocks (each containing two 3x3 convolutions), and concludes with a global average pooling layer and a fully connected layer;  The local learning rate $\eta_l$ is set to 0.1 with rate decay of 0.995. 
\textbf{3) Baselines for comparison:} FedAvg\cite{fl_og}, Scaffold\cite{scaffold}, Feddyn\cite{feddyn} are used for performance comparison.

\subsection{Results and Analysis}
\subsubsection{Robustness against perturbations in different scenarios}
We compare the test accuracy of SMRFL and FedAvg on both Mnist and Cifar-10 datasets under three scenarios (only downlink perturbation, only uplink perturbation, and both uplink and downlink perturbations); results are as depicted in Fig. \ref{exp1}. $(x,y)$ in the legend means uplink perturbation strength of $x$ and downlink perturbation strength of $y$. $0.06$, $0.09$, and $0.12$ are about $8$dB, $11$dB, and $14$dB of perturbations over the model parameters of the simple CNN on Mnist, respectively. There is a noticeable decline in model performance across all scenarios in the presence of noise, highlighting the necessity to investigate robustness against non-malicious perturbations. Specifically, when comparing the impact of different perturbation scenarios, it is evident that downlink perturbations lead to greater degradation in model performance compared to uplink perturbations of equivalent strength. Furthermore, the simultaneous occurrence of both uplink and downlink perturbations exacerbates this effect, resulting in more significant performance losses than either type of perturbation alone. By comparing the first row of sub-figures and the second row, we observe that introducing perturbations has a more pronounced negative impact on the Cifar10. This suggests that more complex data sets may be more susceptible to the effects of perturbations. Notably, SMRFL demonstrates superior performance over FedAvg across various perturbation intensities and scenarios on both datasets. The test accuracy achieved by SMRFL consistently outperforms that of FedAvg, indicating its enhanced robustness. Moreover, SMRFL exhibits faster convergence rates, further underscoring its robustness against perturbations. 
\vspace{-0.1cm}
\subsubsection{Impact of non-IID data}
We further compare the test accuracy of SMRFL with the baseline methods under both IID and non-IID data settings. Mnist is used, and both uplink and downlink perturbations are set to 0.06 in this experiment. Results are presented in table \ref{tab_mnist_perturbation}. Non-IID data can degrade the model accuracy, with a higher degree of non-IID ($\alpha=0.05$) having a more pronounced negative impact on model performance than a lower degree ($\alpha=0.3$). Overall, SMRFL shows superior performance even in the challenging non-IID setting, showcasing its robustness against non-malicious perturbations compared to baseline methods.

\begin{table}[htbp]
	\vspace{-0.3cm} 
	\centering
	\small
	\setlength{\tabcolsep}{3pt} 
	\renewcommand{\arraystretch}{0.8} 
	\caption{Test Accuracy (\%) on MNIST with different perturbation processes and algorithms}
	\label{tab_mnist_perturbation}
	\begin{tabular}{@{}l l p{1.3cm} p{1.3cm} c p{1.3cm}@{}}
		\toprule
		\multirow{2}{*}{\makecell[l]{Perturbed\\process}} & \multirow{2}{*}{Algorithm} & \multicolumn{2}{c}{Non-IID} & & \multirow{2}{*}{IID} \\
		\cmidrule{3-4}
		& & $\alpha=0.05$ & $\alpha=0.3$ & & \\
		\midrule
		\multirow{5}{*}{Uplink} & FedAvg & 90.67 & 97.96 & & 98.56\\
		& Scaffold & 87.31 & 95.76 & & 97.15 \\
		& Feddyn & 86.21 & 96.62 & & 97.09 \\
		& SMRFL & \bf96.20 & \bf98.97 & & \bf99.34 \\
		\midrule
		\multirow{5}{*}{Downlink} & FedAvg & 89.18 & 96.69 & & 97.93 \\
		& Scaffold & 85.46 & 96.86 & & 97.52 \\
		& Feddyn & 86.49 & 96.00 & & 96.98 \\
		& SMRFL & \bf94.94 & \bf98.12 & & \bf99.01 \\
		\midrule
		\multirow{5}{*}{Up\&Down link} & FedAvg & 88.27 & 95.99 & & 96.57 \\
		& Scaffold & 80.31 & 88.73 & & 91.95 \\
		& Feddyn & 84.32 & 95.67 & & 96.81 \\
		& SMRFL & \bf92.70 & \bf97.92 & & \bf98.25 \\
		\bottomrule
	\end{tabular}
	\vspace{-0.4cm}
\end{table}

\begin{figure}[htbp]
	\centering
	\begin{subfloat}[\footnotesize FedAvg]{
			\includegraphics[width=0.22\textwidth]{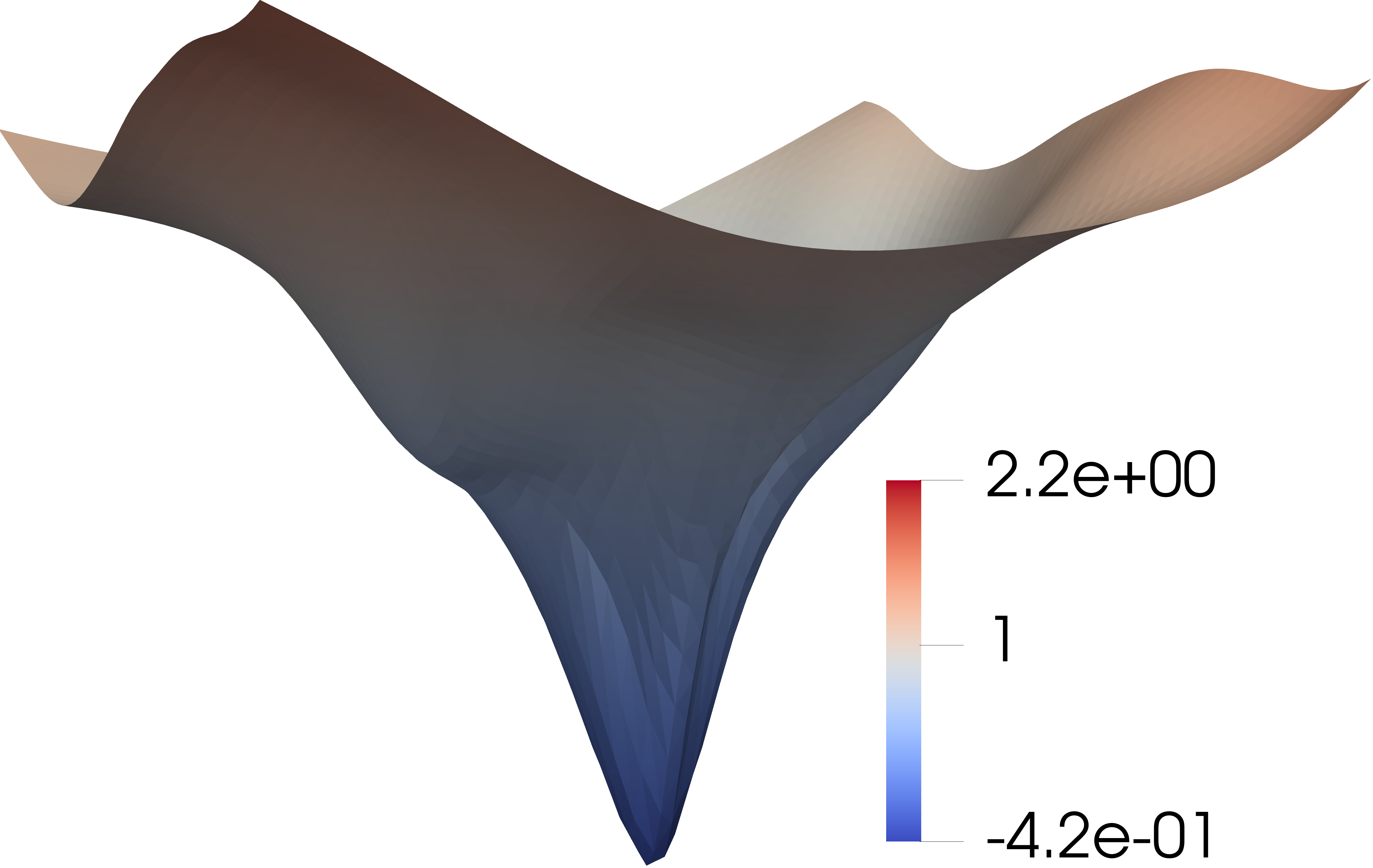}				
			\label{exp2_1}
		}				
	\end{subfloat}
	\vspace{-0.2cm}
	\hspace{-0.3cm}
	\begin{subfloat}[\footnotesize SMRFL $\rho=0.1$]
		{
			\includegraphics[width=0.19\textwidth]{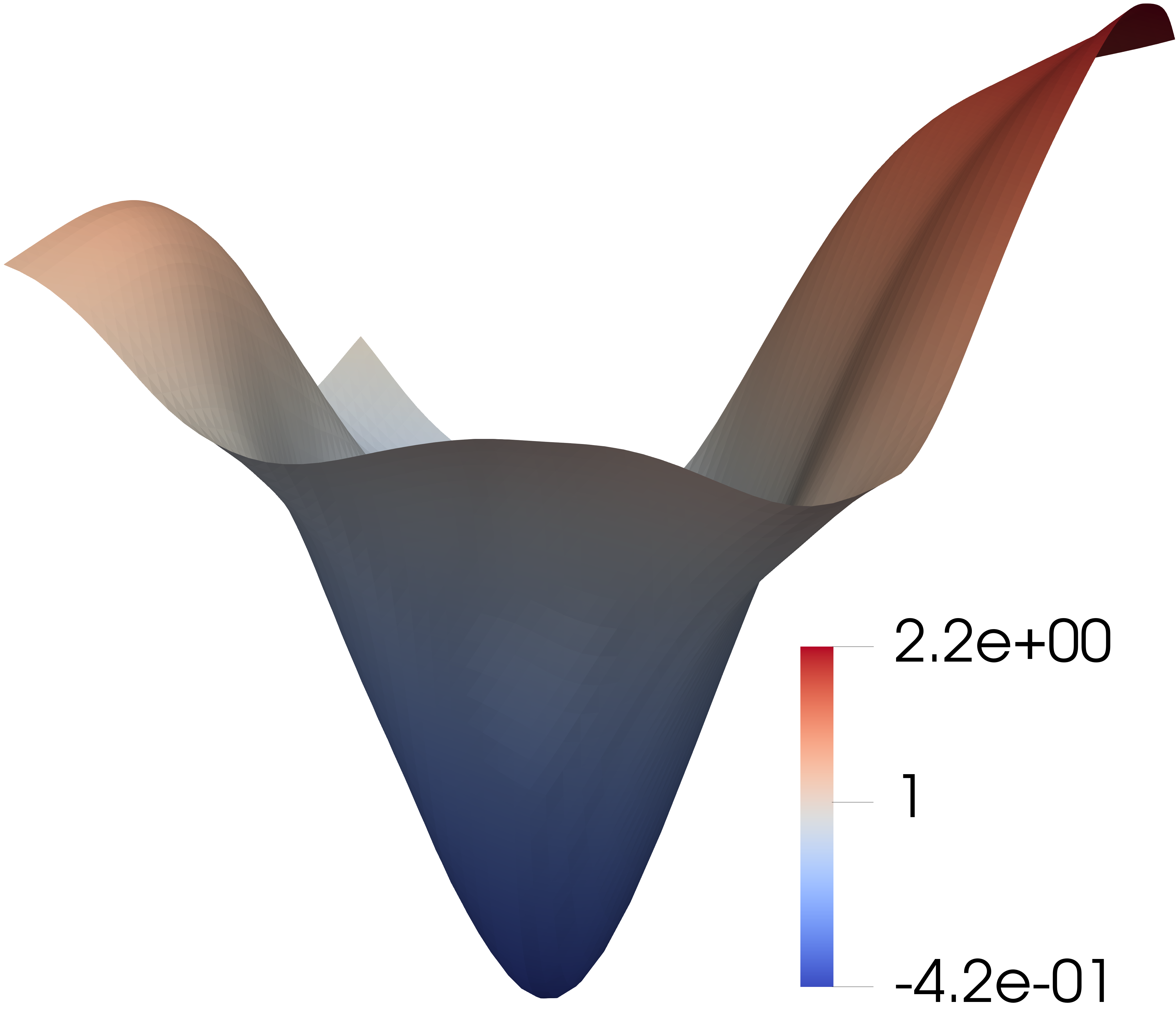}				
			\label{exp2_2}
		}
	\end{subfloat}
	\vspace{0.2cm}
	\begin{subfloat}[\footnotesize SMRFL $\rho=0.5$]{
			\includegraphics[width=0.19\textwidth]{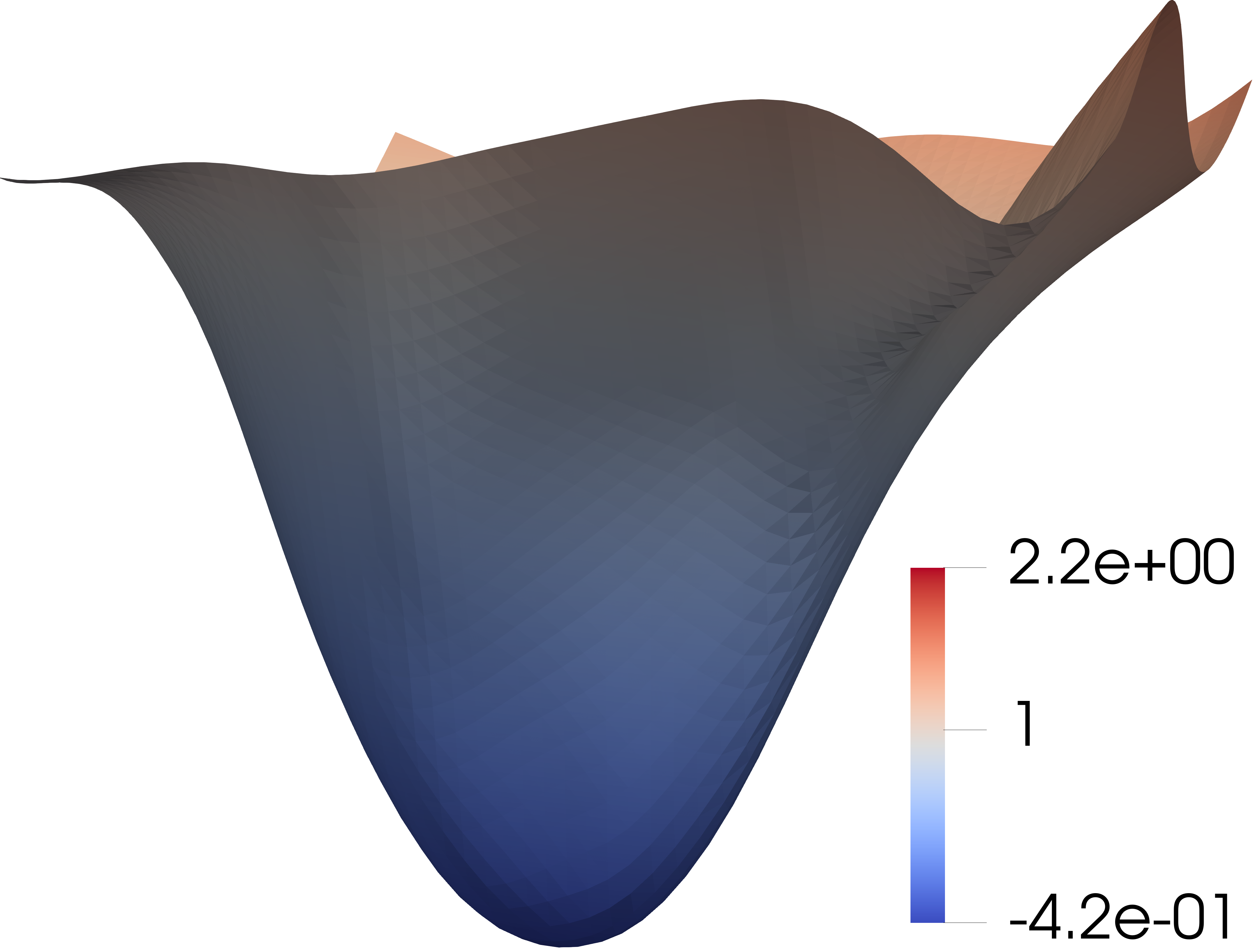}				
			\label{exp2_3}
		}				
	\end{subfloat}
	\hspace{-0.3cm}
	\vspace{0.5cm}
	\begin{subfloat}[\footnotesize SMRFL]{
			\includegraphics[width=0.22\textwidth]{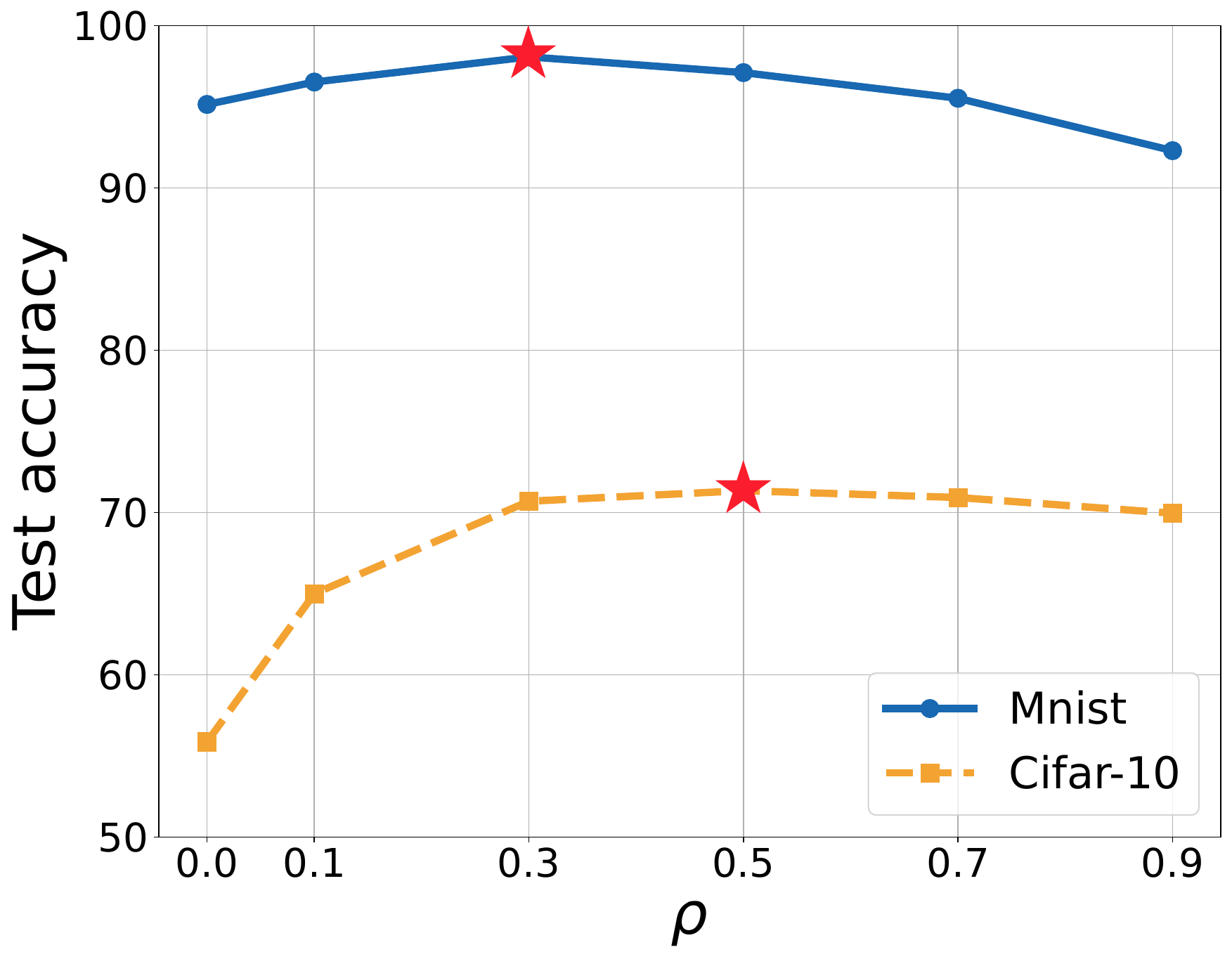}				
			\label{exp2_4}
		}				
	\end{subfloat}
	\caption{Impact of neighborhood radius $\rho$ on the sharpness of loss landscape and test accuracy}
	\label{exp2}
	\vspace{-0.5cm}
\end{figure}

\noindent\subsubsection{Impact of neighborhood radius $\rho$}
Fig. \ref{exp2} illustrates the impact of the neighborhood radius $\rho$ on test accuracy and the sharpness of the loss landscape. Both uplink and downlink perturbations are set to 0.06 in this experiment. The loss landscape is plotted following the method in \cite{visual_loss}. From Fig. \ref{exp2_2}, we can observe that when $\rho$ is set to a small value, SMRFL converges to a sharp minimum as FedAvg in Fig. \ref{exp2_1}. When $\rho$ increases to 0.5 as in Fig. \ref{exp2_3}, it becomes evident that the model converges to a flatter minimum, indicating a more robust model. To investigate the impact of $\rho$ on test accuracy, it is searched from $\{0, 0.1, 0.3, 0.5, 0.7, 0.9\}$ on both MNIST and CIFAR-10 datasets, and results are shown in Fig. \ref{exp2_4}. We find that $\rho = 0.3$ performs better on MNIST, and $\rho = 0.5$ performs better for CIFAR-10. This indicates that the choice of $\rho$ should be carefully considered in practice.

						
\section{Conclusions} 
In this work, we have developed a robust FL framework, namely SMRFL, designed to mitigate the impact of non-malicious perturbations in edge networks. SMRFL addresses the model performance degradation caused by non-malicious perturbations by utilizing SAM, which encourages convergence to flat minima and enhances robustness. Extensive experiments on real-world datasets show that SMRFL significantly improves robustness against non-malicious perturbations, even in the presence of non-IID data.
\vspace{-0.2cm}
\section*{Acknowledgment}
The work of Yingyu Li was supported in part by the National Natural Science Foundation of China under Grant 62301516. The work of Yingyu Li and Yong Xiao was supported in part by the Mobile Information Network National Science and Technology Key Project under grant 2024ZD1300700.

\addtolength{\topmargin}{0.18in}

\bibliography{RFL_RA_conf}

\begin{thebibliography}{10}
\providecommand{\url}[1]{#1}
\csname url@samestyle\endcsname
\providecommand{\newblock}{\relax}
\providecommand{\bibinfo}[2]{#2}
\providecommand{\BIBentrySTDinterwordspacing}{\spaceskip=0pt\relax}
\providecommand{\BIBentryALTinterwordstretchfactor}{4}
\providecommand{\BIBentryALTinterwordspacing}{\spaceskip=\fontdimen2\font plus
\BIBentryALTinterwordstretchfactor\fontdimen3\font minus
  \fontdimen4\font\relax}
\providecommand{\BIBforeignlanguage}[2]{{%
\expandafter\ifx\csname l@#1\endcsname\relax
\typeout{** WARNING: IEEEtran.bst: No hyphenation pattern has been}%
\typeout{** loaded for the language `#1'. Using the pattern for}%
\typeout{** the default language instead.}%
\else
\language=\csname l@#1\endcsname
\fi
#2}}
\providecommand{\BIBdecl}{\relax}
\BIBdecl

\bibitem{XY2024TMCTSFL}
Y.~Xiao \emph{et~al.}, ``Time-sensitive learning for heterogeneous federated
  edge intelligence,'' \emph{IEEE Transactions on Mobile Computing}, vol.~23,
  no.~2, pp. 1382--1400, Feb. 2024.

\bibitem{XY2024TMCTraffSynth}
------, ``Distributed traffic synthesis and classification in edge networks: A
  federated self-supervised learning approach,'' \emph{IEEE Transactions on
  Mobile Computing}, vol.~23, no.~2, pp. 1815--1829, Feb. 2024.

\bibitem{jdz_vtc}
D.~Jin, Y.~Li, and Y.~Xiao, ``Federated generative learning for digital twin
  network modeling,'' in \emph{2024 IEEE 99th Vehicular Technology Conference},
  Singapore, Singapore, Jun. 2024.

\bibitem{robust_qi_2022}
Q.~Qi and X.~Chen, ``Robust design of federated learning for edge-intelligent
  networks,'' \emph{IEEE Trans. Commun.}, vol.~70, no.~7, pp. 4469--4481, Jul.
  2022.

\bibitem{sparse_dp_hu}
R.~Hu, Y.~Guo, and Y.~Gong, ``Federated learning with sparsified model
  perturbation: Improving accuracy under client-level differential privacy,''
  \emph{IEEE Trans. Mob. Comput.}, vol.~23, no.~8, pp. 8242--8255, Aug. 2024.

\bibitem{compu_expensive_1}
J.~So, B.~Güler, and A.~S. Avestimehr, ``Byzantine-resilient secure federated
  learning,'' \emph{IEEE J. Sel. Areas Commun.}, vol.~39, no.~7, pp.
  2168--2181, Jul. 2021.

\bibitem{NEURIPS2021_692baebe}
J.~Sun, A.~Li, L.~DiValentin, A.~Hassanzadeh, Y.~Chen, and H.~Li, ``Fl-wbc:
  Enhancing robustness against model poisoning attacks in federated learning
  from a client perspective,'' in \emph{NIPS}, Virtual, Dec. 2021.

\bibitem{zhao2022fedinv}
B.~Zhao, P.~Sun, T.~Wang, and K.~Jiang, ``Fedinv: Byzantine-robust federated
  learning by inversing local model updates,'' in \emph{AAAI}, Virtual, Feb.
  2022.

\bibitem{qua2022honig}
R.~H{\"o}nig, Y.~Zhao, and R.~Mullins, ``{DA}da{Q}uant: Doubly-adaptive
  quantization for communication-efficient federated learning,'' in
  \emph{ICML}, Baltimore, Maryland, USA, Jul. 2022.

\bibitem{wei2022noise}
X.~Wei and C.~Shen, ``Federated learning over noisy channels: Convergence
  analysis and design examples,'' \emph{IEEE Trans. Cognit. Commun.
  Networking}, vol.~8, no.~2, pp. 1253--1268, Jun. 2022.

\bibitem{foret2020sharpness}
P.~Foret, A.~Kleiner, H.~Mobahi, and B.~Neyshabur, ``Sharpness-aware
  minimization for efficiently improving generalization,'' in \emph{ICLR},
  Virtual, May 2021.

\bibitem{Amiri2022_noise}
M.~M. Amiri, D.~Gündüz, S.~R. Kulkarni, and H.~V. Poor, ``Convergence of
  federated learning over a noisy downlink,'' \emph{IEEE Trans. Wireless
  Commun.}, vol.~21, no.~3, pp. 1422--1437, Mar. 2022.

\bibitem{qua2021shle}
N.~Shlezinger, M.~Chen, Y.~C. Eldar, H.~V. Poor, and S.~Cui, ``Uveqfed:
  Universal vector quantization for federated learning,'' \emph{IEEE Trans.
  Signal Process.}, vol.~69, pp. 500--514, Dec. 2021.

\bibitem{scaffold}
S.~P. Karimireddy, S.~Kale, M.~Mohri, S.~Reddi, S.~Stich, and A.~T. Suresh,
  ``{SCAFFOLD}: Stochastic controlled averaging for federated learning,'' in
  \emph{ICML}, Virtual, Jul. 2020.

\bibitem{fedsam}
Z.~Qu, X.~Li, R.~Duan, Y.~Liu, B.~Tang, and Z.~Lu, ``Generalized federated
  learning via sharpness aware minimization,'' in \emph{ICML}, Baltimore,
  Maryland, USA, Jul. 2022.

\bibitem{fl_og}
B.~McMahan, E.~Moore, D.~Ramage, S.~Hampson, and B.~A. y~Arcas,
  ``Communication-efficient learning of deep networks from decentralized
  data,'' in \emph{AISTATS}, Ft. Lauderdale, FL, USA, Apr. 2017.

\bibitem{feddyn}
D.~A.~E. Acar, Y.~Zhao, R.~Matas, M.~Mattina, P.~Whatmough, and V.~Saligrama,
  ``Federated learning based on dynamic regularization,'' in \emph{ICLR},
  Virtual, May 2021.

\bibitem{visual_loss}
H.~Li, Z.~Xu, G.~Taylor, C.~Studer, and T.~Goldstein, ``Visualizing the loss
  landscape of neural nets,'' in \emph{NIPS}, Montreal Canada, Dec. 2018.

\end{thebibliography}
\bibliographystyle{IEEEtran}

\end{document}